\documentclass[letterpaper, 10 pt, conference]{ieeeconf}  

\IEEEoverridecommandlockouts                              

\usepackage[T1]{fontenc}
\usepackage{caption}
\usepackage[dvipsnames]{xcolor}
\usepackage{graphicx}
\title{\LARGE \bf
Center Feature Fusion: Selective Multi-Sensor Fusion of Center-based Objects
}

\author{Philip Jacobson$^{1}$, Yiyang Zhou$^{1}$, Wei Zhan$^{1}$, Masayoshi Tomizuka$^{1}$ and Ming C. Wu$^{1}$
\thanks{$^{1}$University of California, Berkeley
        {\tt\small \{philip\_jacobson, yiyang.zhou, wzhan, tomizuka\}@berkeley.edu}
        \tt\small wu@eecs.berkeley.edu}%
}

\begin{document}

\maketitle
\thispagestyle{empty}
\pagestyle{empty}

\begin{abstract}

Leveraging multi-modal fusion, especially between camera and LiDAR, has become essential for building accurate and robust 3D object detection systems for autonomous vehicles. Until recently, point decorating approaches, in which point clouds are augmented with camera features, have been the dominant approach in the field. However, these approaches fail to utilize the higher resolution images from cameras. Recent works projecting camera features to the bird's-eye-view (BEV) space for fusion have also been proposed, however they require projecting millions of pixels, most of which only contain background information. In this work, we propose a novel approach Center Feature Fusion (CFF), in which we leverage center-based detection networks in both the camera and LiDAR streams to identify relevant object locations. We then use the center-based detection to identify the locations of pixel features relevant to object locations, a small fraction of the total number in the image. These are then projected and fused in the BEV frame. On the nuScenes dataset, we outperform the LiDAR-only baseline by 4.9\% mAP while fusing up to 100x fewer features than other fusion methods.

\end{abstract}

\section{INTRODUCTION}
3D object detection is a central pillar of the perception system for modern autonomous vehicles. To tackle this problem, most vehicles employ a full suite of sensors, most prominently cameras and LiDARs. LiDAR, the main workhorse in most detection systems, provides low-resolution but accurate spatial and depth information, whereas camera provides high resolution color and texture information. The process of extracting and combining useful complementary information from each of these sensor modalities, known as sensor fusion, is thus of immense importance to building accurate detection models.

The most recent fusion approaches in the literature can be grouped into two main categories. In the first category, raw LiDAR points are decorated with camera features, such as segmentation class labels or CNN features \cite{Vora_2020_CVPR,wang2021pointaugmenting,fusionpainting,autoalign,autoalignv2}. In the second, deep CNN features from both modalities are associated and fused together, often using a learnable attention mechanism \cite{deepfusion,chen2022futr3d,bai2021pointdsc,liu2022bevfusion}. Point decoration, while performing well on many detection benchmarks, only supplements already existing LiDAR points, failing to fully leverage the higher camera resolution. Meanwhile, associating deep features from both modalities requires some mechanism to associate features between the two views, which is a challenging problem in and of itself.

Recently, fusion in the bird's-eye-view representation has emerged as a compelling approach. BEVFusion \cite{liu2022bevfusion} has recently achieved state-of-the-art performance on both the nuScenes \cite{nuscenes2019} and Waymo Open \cite{waymo} 3D object detection challenges. BEV, unlike the camera's perspective view, is an attractive representation for fusion since it preserves the geometry of the scene. However, projecting camera pixels to BEV is much less efficient than projecting 3D points to perspective view because a) we require per-pixel depth estimates to perform the projection and b) the number of camera pixels far outnumbers the number of points in a LiDAR pointcloud. 

In this paper, we propose a novel fusion approach which also leverages the power of the BEV representation. However, we argue that projecting all of the camera features into BEV is wasteful, since the vast majority of pixels simply represent background. Instead, we leverage center-based detector CenterNet \cite{zhou2019objects} to identify objects in perspective view. Since CenterNet infers bounding box characteristics from only a single point, all of the relevant features are agglomerated in a single location per object, making projection to BEV natural and efficient. We therefore set a confidence threshold in the detection heatmap which is used to decide which pixels to project to BEV. After these features are projected, the BEV features pass through a fully-convolutional encoder to help with alignment. Lastly, the aligned features are concatenated with BEV LiDAR features before a CenterPoint regression head calculates the final bounding box locations.

Additionally, the issue of preserving alignment when applying multi-modal data augmentation is essential in fusion work. To enable the training of our model, we apply data augmentation in a novel fashion: first, we apply standard LiDAR augmentations (random flip, scaling, rotation, and translation) \cite{zhu2019class}. Next, we save the random parameters of the augmentation. Once the camera features are projected into global 3D coordinates, we then apply the augmentation on the camera features. This allows us to apply the exact same augmentations to both set of features in an efficient manner.

   \begin{figure*}[thpb]
      \centering
      \includegraphics[scale=0.38]{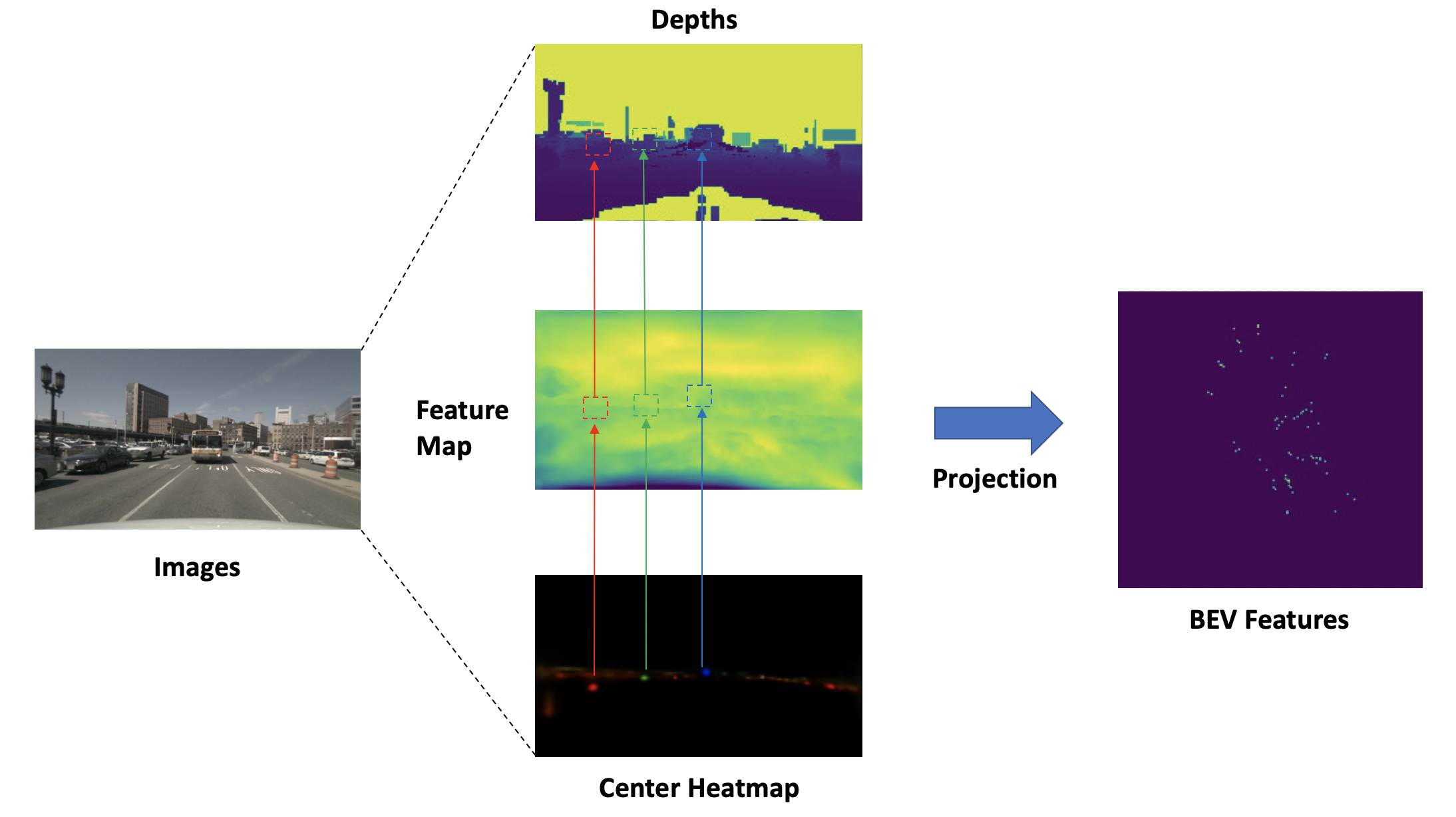}

      \caption{Our proposed selective feature projection. The detection heatmap generated by CenterNet's detection head is used to select interesting regions from the corresponding deep feature map. Dense depth predictions are then used to project the corresponding features to BEV, where they can be fused with LiDAR BEV features.}
      \label{projection}
   \end{figure*}

In summary, our contributions are the following:
\begin{enumerate}
    \item We explore a selective deep feature fusion strategy in the bird's-eye-view domain through projecting camera features based on their confidence score in a detection heatmap.
    \item We propose a novel mechanism for data augmentation that aligns features projected to bird's-eye-view.
    \item We validate our method on the nuScenes dataset, achieving a significant +4.9\% improvement in Mean Average Precision (mAP) and +2.4\% nuScenes Detection Score (NDS) over a LiDAR-only baseline. Our performance is in-line with other fusion methods, even as we fuse a far smaller number of features.
\end{enumerate}

\section{RELATED WORK}
\subsection{LiDAR-based 3D Object Detection}
LiDAR-based 3D object detectors aim to detect objects with a given point cloud, represented as an unordered set of 3D points in space. PointNet and its derivatives \cite{qi2016pointnet,qi2017pointnetplusplus} operate directly on said point set. Voxel-based methods first quantize point clouds into regular 3D voxels before the voxelized representation is processed through the detector stage. VoxelNet \cite{voxelnet} discretizes the 3D space before processing points inside each voxel with a PointNet-like network; voxels are then processed using a dense 3D convolutional neural network. SECOND \cite{second} improves on VoxelNet by introducing efficient sparse convolutions. Pillar-based methods discretize the point cloud into vertical columns, flattening the representation along the height dimension. PIXOR \cite{yang2018pixor} flattens the point cloud into a 2D pseudo-image which is encoded with a 2D CNN. PointPillars \cite{Lang2019PointPillarsFE}, similar to VoxelNet, processes points inside pillars through a PointNet before the encoded pillars are passed through a 2D convolutional network. CenterPoint replaced the standard two-stage detection architecture with an anchor-free approach by utilizing a center-based representation of detected objects, achieving the new state-of-the-art on 3D detection tasks \cite{yin2021center}. 

\subsection{Camera-based 3D Object Detection}
3D object detection using only camera images, although attractive as an alternative to expensive 3D sensors, still lags significantly behind LiDAR-based detection due to the challenge of monocular depth estimation. Pseudo-lidar methods use a pre-trained depth estimation network to transform RGB images into point clouds that are then compatible with standard LiDAR-based detectors \cite{wang2019pseudo,qian2020end}. Another class of detectors transforms camera features into BEV representation. BEVFormer \cite{li2022bevformer} achieves state-of-the-art results by utilizing a temporal cross-attention to calculate interactions between camera features in both the current and previous frame. CenterNet, the 2D counterpart to CenterPoint, similarly uses center-based object detection in perspective view combined with a depth-estimation head to generate 3D bounding box characteristics \cite{zhou2019objects}.

\subsection{Multi-modal Fusion}
Because of the unique advantages and disadvantages of different sensor modalities, leveraging data from each of these modalities is the most promising approach for 3D object detection. Early sensor fusion work focused primarily on proposal-level fusion, i.e. fusing information at the region proposal level. MV3D \cite{mv3d} fuses object proposals in three views (BEV, front view, and perspective view), whereas AVOD fuses them in two views (BEV and perspective view) \cite{ku2018joint}. RoarNet determines geometrically feasible 3D poses from images to generate a set of 3D region proposals \cite{roarnet}. Frustum-PointNet \cite{qi2017frustum} and Frustum-ConvNet \cite{wang2019frustum} lift image proposals into 3D frustums to identify relevant regions of the point cloud. FUTR3D \cite{chen2022futr3d} and TransFusion \cite{bai2021pointdsc} generate object proposals in 3D and refine them using a transformer decoder which attends to 2D features.

Point decoration has been a popular paradigm in sensor fusion for the past few years. PointPainting \cite{Vora_2020_CVPR} proposed projecting the point cloud onto the outputs of 2D semantic segmentation networks to augment each point with the class label of the corresponding 2D pixel. FusionPainting \cite{fusionpainting} fuses segmentation labels from both 2D and 3D segmentations to generate more accurate labels for the point cloud. PointAugmenting \cite{wang2021pointaugmenting} instead annotates points with deep CNN features while processing the LiDAR and camera features through separate streams in the CNN backbone. AutoAlign \cite{autoalign} and AutoAlignV2 \cite{autoalignv2} use a learnable cross-attention module to associate pixel-level features with voxels, instead of the hard association of projection using a camera matrix. MVP \cite{yin2021multimodal} takes a different input decorating approach, projecting the point cloud onto a segmentation map to generate virtual LiDAR points inside each object to densify the point cloud.

Most recently, mid-level fusion in which fusion is done at the deep-feature level has become popular. Continuous Fusion \cite{contfuse2018} shares information between the 2D and 3D backbones at all layers of the neural network. DeepFusion \cite{deepfusion} uses a learnable attention layer to associate CNN features between different sensor modalities. BEVFusion \cite{liu2022bevfusion} uses the approach from LSS \cite{philion2020lift} to generate per-pixel depth predictions to project the full set of camera features to BEV, where after undergoing a pooling operation, they are concatenated with LiDAR BEV features.

   \begin{figure*}[thpb]
      \centering
      \includegraphics[scale=0.45]{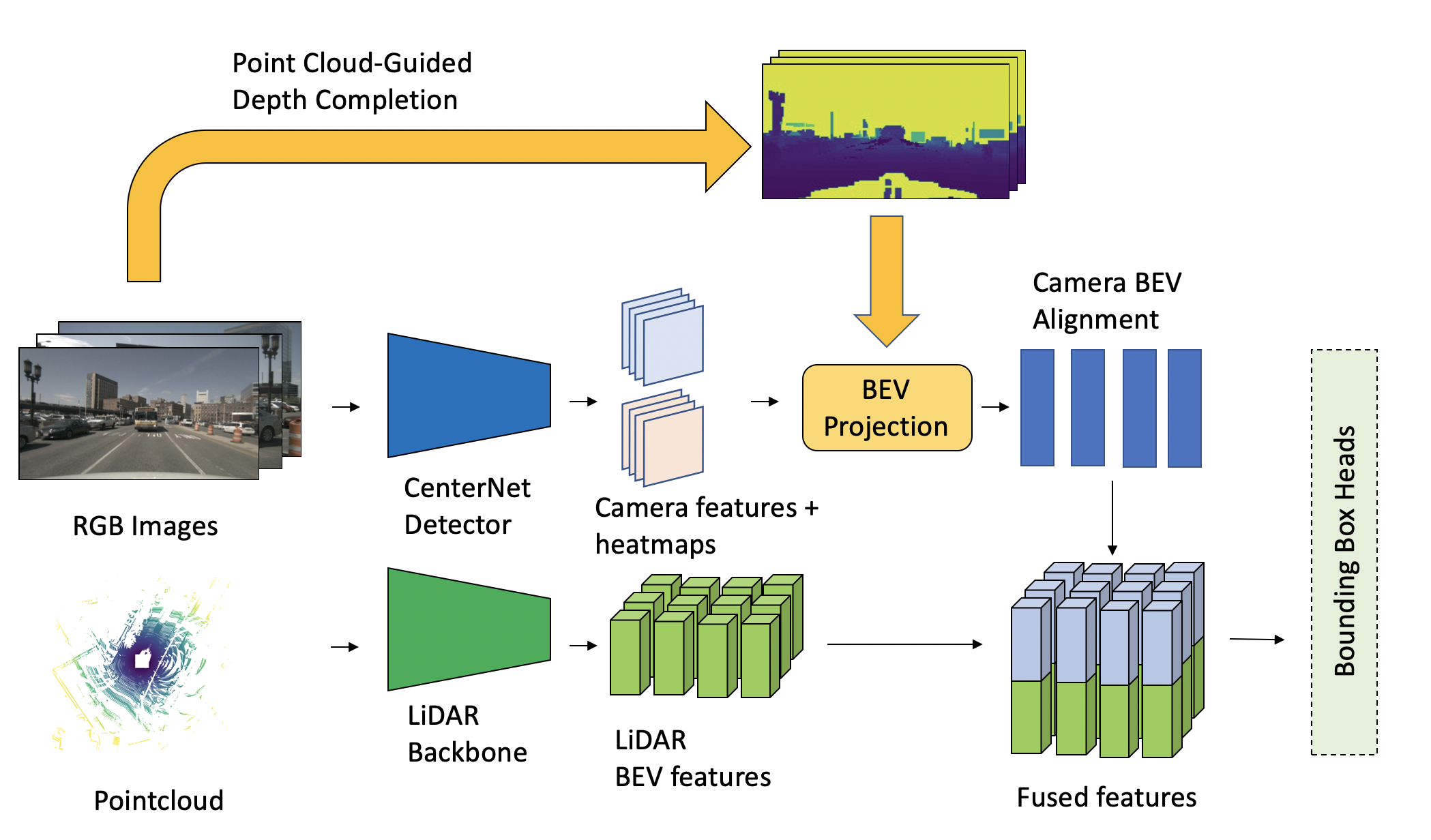}
      \caption{Summary of our proposed Center Feature Fusion model. The pointcloud and camera images are first encoded using a VoxelNet and DLA34-CenterNet, respectively. Our selective projection is performed on the camera features to project them to BEV where they are then fed into a small 2D CNN to help with feature alignment. The multi-modal BEV features are then concatenated and passed into CenterPoint's bounding box heads.}
      \label{Architecture}
   \end{figure*}

\section{CENTER FEATURE FUSION}

This section introduces our proposed method, Center Feature Fusion. Our model uses CenterPoint \cite{yin2021center} and CenterNet \cite{zhou2019objects} as 3D and 2D feature encoders, respectively, to generate features for fusion. The CenterNet encoder additionally generates detection heatmaps in the perspective view, which we use to select the relevant features for projection. We generate per-pixel depth estimates and then use these to project the selected camera features to BEV. The camera BEV features are encoded using a further 2D convolutional network before concatenation with LiDAR features. The concatenated features are then passed to CenterPoint's various bounding box heads. A detailed diagram illustrating this method is shown in Fig. \ref{Architecture}. Additionally, we introduce a novel data augmentation scheme, ProjectionAlign, to facilitate the training of the fusion model. To ensure alignment between 2D and 3D features, we apply data augmentation to the projected 2D camera features, and not the images at the input, so that the same exact augmentations can be applied to both set of features.

\subsection{Camera Feature Representation}
Given the high resolution of cameras relative to LiDARs, the 2D feature extractors of most fusion models generate significantly more features than those of the LiDAR backbone. However, utilizing all of these features can represent a significant bottleneck; projecting potentially millions of pixels to BEV is a computationally intensive task, while allowing each LiDAR feature to attend to all camera features greatly increases model complexity. We observe that the vast majority of camera features are irrelevant for object detection, only representing the background. CenterNet takes this to the extreme by relying on only the features located at a single center pixel per object to characterize it. Thus, we argue for an approach which selectively chooses relevant features to fuse, utilizing CenterNet's 2D detection heatmap.

\subsection{Camera Detection Pipeline}
Before fusion, camera images are fed into a pre-trained CenterNet object detector. CenterNet's heatmap head generates a keypoint heatmap $\hat{Y} \in [0,1]^{\frac{w}{s} \times \frac{h}{s} \times K}$, where $w$ and $h$ are the input height and width, respectively, $s$ is the downsampling factor of the output, and $K$ is the number of classes. Each local peak in the heatmap (i.e. a point with value larger than its eight neighbors) corresponds to a detected object, and the bounding box characteristics are then inferred from the features located at that point. Because of this, we argue that all of the information relevant for detection and bounding box regression are located at these central pixels. We selectively fetch these points by choosing a heatmap threshold $\in [0,1]$, pixels below the threshold are disregarded. A lower threshold, while fetching more features per object as well as features for less confident detections, increases latency. For each pixel location above this threshold, the corresponding location in the output feature map of the DLA34 \cite{Yu_2018_CVPR} CenterNet backbone is chosen as the camera feature. Fig. \ref{projection} illustrates our selective projection method.

In order to project the selected features from perspective view to BEV, we require per-pixel depth estimates. While BEVFusion utilizes the approach in LSS of predicting a per-pixel depth distribution \cite{liu2022bevfusion}, this requires training a separate view transformer. Instead, we project LiDAR points to the camera frame and then perform depth completion. MVP utilizes a nearest-neighbor based classifier for depth completion to generate the virtual points, the simplest approach to depth completion \cite{yin2021multimodal}. Many deep-learning approaches exist for depth-completion \cite{jaritz2018,Gansbeke2019,Eldesokey2019,Xu_2019_ICCV}, however these require separate training and are computationally expensive. Instead, we use IP-Basic \cite{ku2018defense}, a fast, lightweight and efficient classical depth completion algorithm which is significantly more accurate than a nearest-neighbor classifier. To prevent objects in the background outside of the LiDAR range from being projected into the foreground, we only perform depth completion on points within interpolation range, and set all other depths outside of the detection range. Using the per-pixel depths estimates, we project the previously selected perspective view features to the 3D world coordinate system. Features are then aggregated inside the cells of the BEV grid, and a max pooling operation is performed on all of the features within a cell to produce the final BEV features.

\subsection{Fusion in BEV Frame}
Once the camera features are organized into the 2D BEV grid, we further encode them using a small 2D convolutional neural network, as in PointAugmenting and BEVFusion. This extra convolutional stream allows the model to compensate for small misalignments from the inherent inaccuracies in the depth estimation process. After this additional refinement, the camera BEV features are concatenated with the LiDAR features, where they then undergo a shared convolutional block before being broadcast to CenterPoint's regression heads.

\subsection{ProjectionAlign Augmentation}
\begin{figure}[thpb]
      \centering
      \includegraphics[scale=0.45]{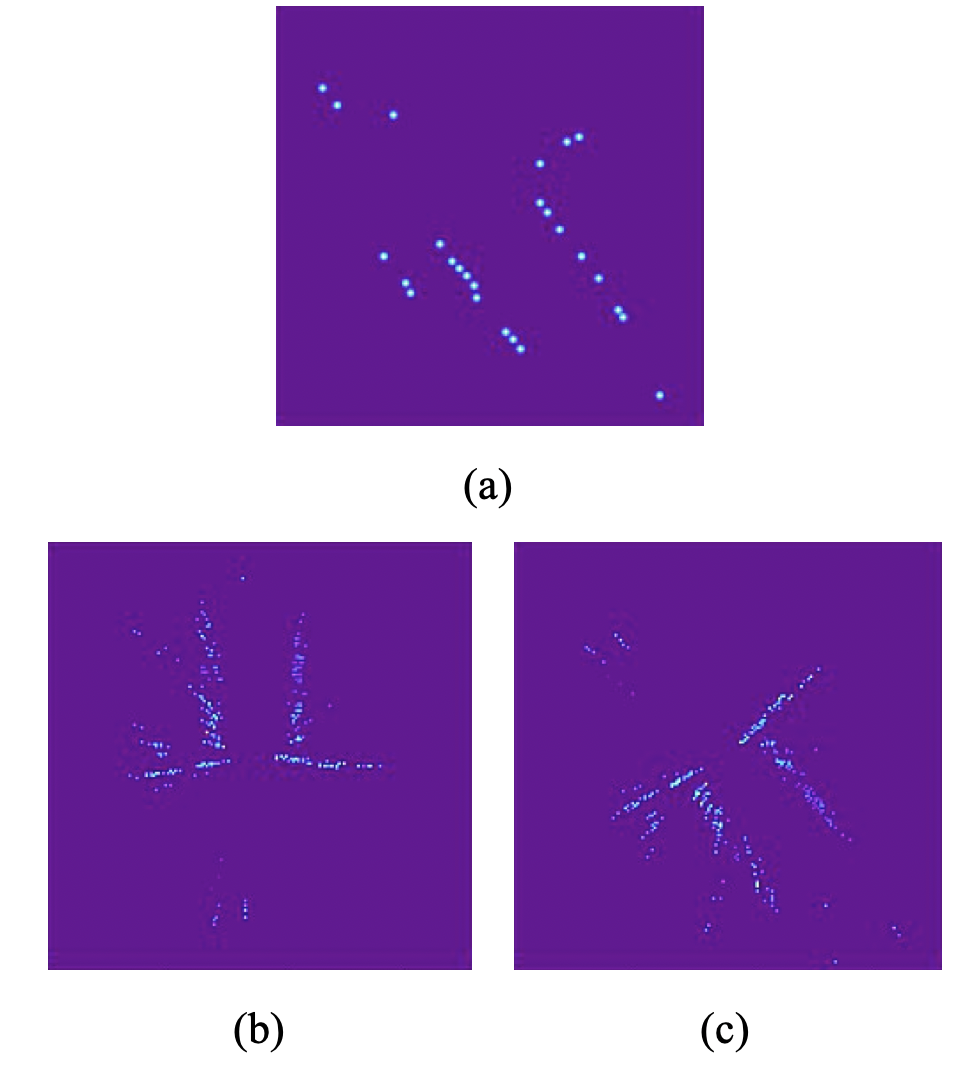}
      \caption{Demonstration of ProjectionAlign. a) GT detections in BEV. b) Unaligned projected camera features. c) Aligned camera features. Much stronger location correspondence exists between aligned features and ground truth object locations vs. unaligned features.}
      \label{alignment}
   \end{figure}

Strong data augmentations are essential during training for most current 3D detection benchmarks to achieve state-of-the-art results. In sensor fusion applications, data augmentation poses a challenge due to the fundamentally different approaches applied in 2D (e.g. random cropping, random flipping) and 3D (random rotation, scaling). Applying these augmentations naively destroys the consistency between fused points and degrades performance. DeepFusion addresses this problem with InverseAug, in which they save the 3D augmentation parameters, reverse them at the fusion stage, and then calculate the feature's corresponding coordinates in the 2D frame. By using BEV as our fusion representation, we are able to propose an even simpler algorithm we call ProjectionAlign. Because we project camera features to BEV, we generate a pseudo-3D point cloud, which can be augmented in the same manner as the LiDAR point cloud. Thus, we simply apply augmentations to the point cloud, save the exact parameters of the augmentation, and apply them to the camera features once they are converted to the same 3D world coordinate system. We don't apply any geometric augmentations to the camera images directly to preserve the exact alignment between the two modalities. A visualization of our alignment scheme is shown in Fig. \ref{alignment}.

\section{EXPERIMENTS}
We evaluate CFF on the nuScenes dataset \cite{nuscenes2019}, a large-scale autonomous driving benchmark. nuScenes consists of 1000 scenes, split into 700 for training, 150 for validation, and 150 for testing. Object annotations are included for every tenth frame in the dataset, resulting in 28130 frames for training, 6019 frames for validation, and 6008 frames for testing. Each frame consists of a 32-beam LiDAR scan and six monocular camera images of resolution 1600 x 900 covering the full 360-degree field-of-view. Annotations come from a set of 10 classes. Detection performance is measured using two main metrics: Mean Average Precision (mAP), and the nuScenes Detection Score (NDS).

\begin{table*}[h]
\begin{center}
\resizebox{\textwidth}{!}{%
\begin{tabular}{c|c|c|c c c c c c c c c c}
\hline
 & mAP & NDS & Car & Truck & C.V. & Bus & Trailer & Barrier & Motor. &  Bike & Ped. & T.C.\\
\hline
\hline
CenterPoint \cite{yin2021center} & 59.6 &  66.8 & 85.5 & 58.6 & 17.1 & 71.5 & 37.2 & 68.5 & 58.9 & 43.3 & 85.1 & 69.7\\
\hline
Ours & \textbf{64.5} & \textbf{69.2} & \textbf{85.5} & \textbf{59.1} & \textbf{22.5} & \textbf{73.4} & \textbf{43.2} & \textbf{68.8} & \textbf{72.8} & \textbf{62.1} & \textbf{86.4} & \textbf{70.6}\\
\hline
Improvement & \color{Green} +4.9 & \color{Green} +2.4 & \color{Green} +0.0 & \color{Green}+0.5 & \color{Green} +5.4 & \color{Green} +1.9 & \color{Green} +6.0 & \color{Green} +0.3 & \color{Green} +13.9 & \color{Green} +18.8 & \color{Green} +1.3 & \color{Green} +0.9\\
\hline
\end{tabular}}
\caption{Performance comparison of CenterPoint and our method on nuScenes validation set. We report the total Mean Average Precision (mAP) and nuScenes Detection Score (NDS), and per-class mAP. C.V., Motor., Ped., and T.C. are short for Construction Vehicle, Motorcyle, Pedestrian, and Traffic Cone, respectively. }
\label{improvement_comparison}
\end{center}
\end{table*}

\begin{table*}[h]
\begin{center}
\resizebox{\textwidth}{!}{%
\begin{tabular}{c|c|c|c|c c c c c c c c c c}
\hline
 & \# of fused features & mAP & NDS & Car & Truck & C.V. & Bus & Trailer & Barrier & Motor. &  Bike & Ped. & T.C.\\
\hline
\hline
CenterPoint \cite{yin2021center} & - & 60.3 & 67.3 & 85.2 & 53.5 & 20.0 & 63.6 & 56.0 & 71.1 & 59.5 & 30.7 & 84.6  & 78.4 \\
\hline
PointAugmenting \cite{wang2021pointaugmenting} & $\sim$200000 & 66.8 & 71.0 & 87.5 & 57.3 & 28.0 & 65.2 & 60.7 & 72.6 & 74.3 & 50.9 & 87.9 & 83.6 \\
\hline
TransFusion \cite{bai2021pointdsc} & $\sim$100000 & 68.9 & 71.7 & 87.1 & 60.0 & 33.1 & 68.3 & 60.8 & 78.1 & 73.6 & 52.9 & 88.4 & 86.7 \\
\hline
BEVFusion \cite{liu2022bevfusion} & $\sim$100000 & \textbf{70.2} & \textbf{72.9} & 88.6 & 60.1 & 39.3 & 69.8 & 63.8 & 80.0 & 74.1 & 51.0 & 89.2 & 86.5 \\
\hline
 Ours & \textbf{$\sim$1000} & 65.2 & 69.9 & 84.7 & 55.4 & 26.0 & 66.2 & 59.2 & 74.3 & 72.6 & 50.7 & 85.7 & 77.5\\
 \hline
\end{tabular}}
\caption{Comparison with state-of-the-art fusion methods and CenterPoint on nuScenes test set. Our method achieves a similar increase in performance to other fusion approaches, even as we use only approximately $1/100$ the number of features for fusion.}
\label{test_set_comparison}
\end{center}
\end{table*}

\subsection{Implementation Details}
We implement CFF with the open-source code of CenterPoint and CenterNet. The CenterNet model uses a DLA34 backbone which is pre-trained separately on the nuScenes dataset. Our CenterPoint model is built with a VoxelNet backbone. Following CenterPoint, we use a detection range of $[-54m,54m]$ for the $X$ and $Y$ axis, and a range of $[-5m,3m]$ for the $Z$ axis. We use a voxel size of $(0.075m,0.075m,0.2m)$. We perform the standard LiDAR data augmentations of random flipping along the $X$ and $Y$ axis, global random scaling, global random rotation, and global random translation. We use the adamW optimizer \cite{adamw} with a one-cycle learning rate policy, a max learning rate of $1e-3$, weight decay of 0.1 and momentum between 0.85 and 0.95. We adopt the class-balanced sampling and class-grouped training proposed in CBGS \cite{zhu2019class}. We train our model using a batch size of 12 on 3 Nvidia A6000 GPUs.

\begin{table}[h]
\begin{center}
{\normalsize %
\begin{tabular}{c|c|c}
\hline
Augmentation Strategy & mAP & NDS \\
\hline
\hline
No Augmentation & 60.4 & 66.6 \\
+ LiDAR Only & 63.2 & 68.2 \\
+ ProjectionAlign & \textbf{64.5} & \textbf{69.2} \\
\hline
\end{tabular}}
\caption{Ablation study of ProjectionAlign. Results shown for nuScenes validation set. }
\label{projectionalign}
\end{center}
\end{table}

\subsection{Main Results}
\begin{figure*}[thpb]
      \centering
      \includegraphics[scale=0.5]{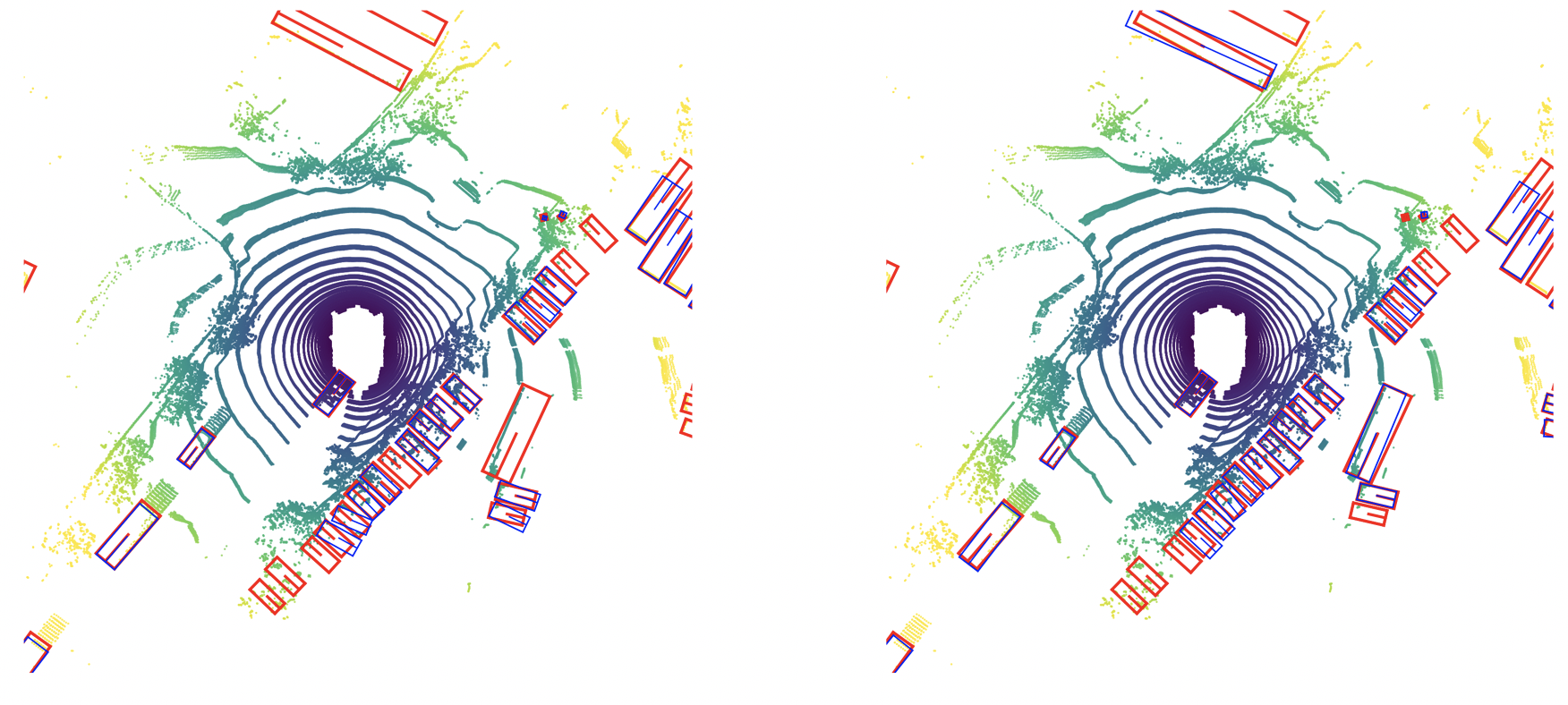}
      \caption{Qualitative Comparison of detection results from CenterPoint (left) and Center Feature Fusion (right). Red boxes indicate ground-truth objects, while blue indicate model predictions. Our approach detects several cars and trucks that CenterPoint misses, as well as generating more accurate boxes for a number of objects.}
      \label{bboxes}
   \end{figure*}

First we compare the improvement from our proposed fusion approach against the CenterPoint LiDAR-only baseline. We display this comparison on the nuScenes validation set in Table \ref{improvement_comparison}. Our method achieves a $+4.9\%$ improvement in mAP and a $+2.4\%$ improvement in NDS overall against CenterPoint, with improvements in mAP seen across all of the classes. Fig \ref{bboxes} visualizes our results qualitatively for an example frame in the nuScenes validation set.

Table \ref{test_set_comparison} compares the performance of our method with other state-of-the-art fusion approaches on the nuScenes test set. In addition to mAP and NDS, we also compare an approximate number of fused features per-frame. PointAugmenting \cite{wang2021pointaugmenting} (and other related point-decoration methods) augment each point in the point cloud, which is, on average, around 200000 points in nuScenes after concatenation with non-key frames. TransFusion \cite{bai2021pointdsc} calculates attention between LiDAR BEV features and CenterNet \cite{zhou2019objects} camera features in perspective view, and so the fused features are aggregated across all more than 100000 pixels in the CenterNet output feature maps. BEVFusion \cite{liu2022bevfusion} projects and fuses all camera features encoded by a Swin-T backbone, amounting to almost 150000 total pixels. In comparison, through our selective fusion approach, we toss out the majority of camera features and fetch only around 1000 features to project and fuse, more than $100\times$ fewer than the aforementioned fusion methods. Despite this, we achieve comparable improvements in performance over the CenterPoint baseline. We outperform PointAugmenting on a handful of classes (e.g. bus, barrier) even without the test-time-augmentation PointAugmenting employs. On the more fine-grained classes of motorcyclists and bicyclist, which rely heavily on semantic information for detection, we achieve performance in-line with all of the other methods.

\begin{table}[h]
\begin{center}
{\normalsize %
\begin{tabular}{c|c|c|c}
\hline
Threshold & \# of projected pixels &  Latency & mAP \\
\hline
\hline
0.5 & 110 & 57 ms & 60.8\\
0.1 & 1442 & 74 ms & 64.5\\
0.05 & 2855 & 98 ms & 64.8\\
0.01 & 20077 & 558 ms & 63.5\\
0.0 & 134400 & 3690 ms & -\\
\hline
\end{tabular}}
\caption{Comparison of average projection latency, number of pixels projected from perspective view to BEV and resulting mAP for various detection heatmap thresholds on nuScenes dataset. Projecting all camera pixels presents a significant processing bottleneck. We omit the mAP for a threshold of 0.0 as the latency makes training this model infeasible.}
\label{heatmapthreshold}
\end{center}
\end{table}

\subsection{Ablation Studies}
\textbf{Augmentation Strategy:} We validate the effectiveness of our ProjectionAlign data augmentation by comparing its performance against no augmentation, augmentation only the standard LiDAR augmentations are applied (i.e. camera and LiDAR features are misaligned), and our augmentation strategy. The results are summarized in Table \ref{projectionalign}. Without any sort of data augmentation, our fusion model quickly runs into overfitting and improvement over the LiDAR-only baseline is minimal at +0.8 mAP. Applying the naive augmentation strategy improves our model by significant amount of +2.8 mAP. By introducing our aligned augmentation strategy, the model performance is further improved  by +1.2 mAP. This confirms the effectiveness of our augmentation scheme and the general importance of aligning fused features, as was shown in \cite{deepfusion}. 

\textbf{Heatmap Thresholding:} To illustrate the computational efficiency gained from our selective projection strategy, we perform experiments varying the heatmap threshold used to determine which camera features should be projected to BEV. These are summarized in Table \ref{heatmapthreshold}. A threshold of 0, meaning every pixel across all 6 cameras is projected, requires on average more than 3.5 seconds on a single NVIDIA A6000 GPU. Given that most of the pixels are identified by the object detector as background, even a small threshold of 0.01 reduces the number of projected pixels and the required latency by more than $6\times$. We find that improvement in mAP levels off at a threshold of 0.1, which results in fusing almost $1/100$ of the total camera features available, illustrating the redundancy of the vast majority of available features. Thus, we use the threshold of 0.1 in all of the experiments presented in this paper.

\section{CONCLUSION}
In this paper, we present a novel fusion approach, Center Feature Fusion, and an associated novel data augmentation scheme, ProjectionAlign. We demonstrate that working in the BEV feature space, fusing only a small number of camera features significantly improves performance of the LiDAR-only baseline, and CFF is comparable in performance to other fusion methods which requires fusing and correlating many times more features. These results show that a more selective approach in choosing which deep features to fuse may allow for improvements in processing speed and computational efficiency without significantly degrading detection accuracy.





\section*{ACKNOWLEDGMENT}
We would like to thank Aaron Drumm for his helpful discussions during the writing of the paper. Philip Jacobson is supported by a National Defense Science and Engineering Graduate Fellowship.

\bibliographystyle{IEEEtran}
\bibliography{report}

\end{document}